\documentclass[runningheads]{llncs}
\usepackage{graphicx}
\usepackage{appendix}

\usepackage{tikz}
\usepackage{comment}
\usepackage{amsmath,amssymb} 
\usepackage{color}

\usepackage{multirow}
\usepackage{wrapfig}
\usepackage{booktabs}

\begin{document}
\pagestyle{headings}
\mainmatter
\def\ECCVSubNumber{5115}  

\title{LabelEnc: A New Intermediate Supervision Method for Object Detection} 

\titlerunning{LabelEnc: A New Intermediate Supervision Method for Object Detection}
%

\newcommand*\samethanks[1][\value{footnote}]{\footnotemark[#1]}
\author{Miao Hao\inst{1}\thanks{Equal contribution. This work is done during Miao Hao and Yitao Liu’s internship at MEGVII Technology and is supported by The National Key Research and Development Program of China (No. 2017YFA0700800) and Beijing Academy of Artificial Intelligence (BAAI).} \and
Yitao Liu\inst{2}\samethanks \and
Xiangyu Zhang\inst{3}\thanks{Corresponding author.} \and
Jian Sun \inst{3}
}

\authorrunning{M. Hao, Y. Liu, X. Zhang and J. Sun}
%
\institute{Beijing University of Posts and Telecommunications\and
Tongji University \and
MEGVII Technology \\
\email{haomiao@bupt.edu.cn, } \email{att@tongji.edu.cn} \\ \email{\{zhangxiangyu,sunjian\}@megvii.com}
}

\maketitle

\begin{abstract}
In this paper we propose a new intermediate supervision method, named LabelEnc, to boost the training of object detection systems. The key idea is to introduce a novel label encoding function, mapping the ground-truth labels into latent embedding, acting as an auxiliary intermediate supervision to the detection backbone during training. Our approach mainly involves a two-step training procedure. First, we optimize the label encoding function via an AutoEncoder defined in the label space, approximating the ``desired'' intermediate representations for the target object detector. Second, taking advantage of the learned label encoding function, we introduce a new auxiliary loss attached to the detection backbones, thus benefiting the performance of the derived detector. Experiments show our method improves a variety of detection systems by around 2\% on COCO dataset, no matter one-stage or two-stage frameworks. Moreover, the auxiliary structures only exist during training, i.e. it is completely cost-free in inference time. Code is available at: \url{https://github.com/megvii-model/LabelEnc}
\keywords{Object Detection, Auxiliary Supervision, AutoEncoder}
\end{abstract}

\section{Introduction}
\label{sec:intro}

Object detection is one of the fundamental problems in computer vision. In deep learning era, modern object detection networks \cite{fasterrcnn,girshick2015fast,fpn,retinanet,maskrcnn,fcos,redmon2016you,redmon2017yolo9000} are composed of two main components: one is the \emph{backbone} part $f(\cdot; \theta_f)$, which generates the intermediate embedding from each image; the other part is the \emph{detection head} $d(\cdot; \theta_d)$, to extract instance information (i.e. class label as well as the corresponding bounding box) from the intermediate representation. To learn the parameters $\theta_f$ and $\theta_d$, earlier work like \cite{he2015spatial} proposes to optimize them separately on different datasets respectively. However, most of recent state-of-the-art detection frameworks \cite{girshick2015fast,fasterrcnn,fpn,redmon2016you,liu2016ssd} suggest \textbf{joint optimization}  of the backbones and detection heads for simpler pipeline and better performance, formulated as follows: 
\begin{equation}
	\theta_f^*, \theta_d^* = \mathop{\arg\min}_{\theta_f, \theta_d} {}\mathbb{E}_{(I, y)\sim\mathcal{D}} \quad
	\mathcal{L}_{det}\left(d(f(I; \theta_f);\theta_d), y \right),
\label{eq:det}
\end{equation}
where $(I, y)$ stands for a pair of image and ground-truth label; $\mathcal{D}$ is the dataset distribution; and $\mathcal{L}_{det}(\cdot, \cdot)$ represents the detection loss, which is usually composed of classification terms and bounding-box regression terms \cite{fasterrcnn}. 


Typically, the backbone part $f(\cdot; \theta_f)$ contains too many parameters, thus may be nontrivial or very costly to be directly optimized in the detection dataset \cite{zhu2019scratchdet,shen2017dsod,li2018detnet,he2019rethinking}. A common practice is to introduce \emph{pretraining}, for instance, initializing $\theta_f$ in Eq.~\ref{eq:det} with \emph{ImageNet} pretrained \cite{fasterrcnn,redmon2016you,redmon2017yolo9000,fpn,liu2016ssd,retinanet} or self-supervised \cite{chen2020simple,he2019momentum} weights. Though such \emph{pretraining-then-finetuning}
paradigm has been demonstrated to achieve state-of-the-art performances \cite{maskrcnn,cai2018cascade}, however, we find that only pretraining backbone weights $\theta_f$ may be suboptimal for the optimization. Since the weights in detection head $\theta_d$ are still randomly initialized, during training, gradient passed from the detection head to the backbone could be very noisy, especially in the very beginning. The noisy gradient may significantly harm the pretrained weights, causing slower convergence or poorer performance. Actually, such degradation has been observed in many codebases and a few workarounds are also proposed. For example, a well-known workaround is to freeze a few weight layers in the backbone during finetuning to avoid unstable optimization \cite{fasterrcnn,girshick2015fast}; however, it seems still insufficient to fully address the issue. 


In this paper, we propose to deal with the problem from a new direction -- introducing an auxiliary intermediate supervision \emph{directly} to the backbone. The key motivation is, if we can provide a feasible supervision in the training phase,  the backbone part could be effectively optimized even before the detection head converges. We formulate our method as follows:
\begin{equation}
\theta_f^*, \theta_d^* = \mathop{\arg\min}_{\theta_f, \theta_d} {}\mathbb{E}_{(I, y)\sim\mathcal{D}} \quad
\mathcal{L}_{det}\left(d(f(I; \theta_f);\theta_d), y \right)
+ \lambda \mathcal{R}(f(I; \theta_f), y), 
\label{eq:ours}
\end{equation}
where $\mathcal{R}(f(I; \theta_f), y)$ means the auxiliary loss attached to the outputs of the backbone, which is \emph{independent} to the detection head $d(\cdot, \theta_d)$ thus not affected by the latter's convergence progress. $\lambda$ is the balanced coefficient. 


The core of our approach thus includes the design of $\mathcal{R}(\cdot, \cdot)$. Intuitively, the auxiliary supervision aims to minimize the distance between latent feature representation and some ``ideal'' embedding of the corresponding training sample. However, how to define and calculate the desired representation? Some previous works, especially \emph{Knowledge Distillation} \cite{knowledgedistillation} methods, suggest acquiring the intermediate supervision from more powerful \emph{teacher models}; nevertheless, whose representations are not guaranteed to be optimal. Instead in this work, for the first time, we point that the \emph{inverse} of the \emph{underlying optimal} detection head (i.e. $d^{-1}(y; \theta_d^*)$) could be the feasible embedding, which traces the ground-truth label $y$ back to the corresponding latent feature. More discussion will be referred in Sec.~\ref{sec:method}.

Motivated by the analyses, in our proposed method \textbf{LabelEnc}, we introduce a novel \emph{label encoding function} to realize $\mathcal{R}(\cdot,\cdot)$ in Eq.~\ref{eq:ours}, which maps the ground-truth labels into the latent embedding space thus providing an auxiliary intermediate supervision to the detector's training. Label encoding function is designed to approximate $d^{-1}(y; \theta_d^*)$, since the \emph{underlying optimal} parameters $\theta_d^*$ and the ``\emph{inverse form}'' $d^{-1}(\cdot)$ in the latter's formulation are nontrivial to be directly derived in mathematics. Thus our method in general involves a \emph{two-step} training pipeline. \textbf{First}, to learn the \emph{label encoding function}, we train an \emph{AutoEncoder} architecture, embedding the ground-truth labels into the latent space according to the (approximated) optimal detection head. \textbf{Second}, with the help of the learned label encoding function, we optimize Eq.~\ref{eq:ours} under the auxiliary supervision $\mathcal{R}(f(I; \theta_f), y)$; in addition, initial weights in the detection head $\theta_d$ can also inherit from the AutoEncoder instead of random values for more stable optimization.  

We evaluate our method on various object detection systems. Under different backbones (e.g. \emph{ResNet}-50, \emph{ResNet}-101 \cite{resnet} and \emph{Deformable Convolutional Networks} \cite{dai2017deformable,dcnv2}) or detection frameworks (e.g. \emph{RetinaNet} \cite{retinanet}, \emph{FCOS} \cite{fcos} and \emph{FPN} \cite{fpn}), on each of them our training pipeline achieves significant performance gains consistently, e.g. $\sim$\textbf{2\%} improvements on \emph{COCO} \cite{coco} dataset. More importantly, our method is completely cost-free in inference time, as the auxiliary structures only exist during training. Please refer to Sec.~\ref{sec:exp} for detailed results.

In conclusion, the major contributions of our paper are as follows:

\begin{itemize}
	\item We propose a new auxiliary intermediate supervision method named \textbf{LabelEnc},  to boost the training of object detection systems. With the novel \emph{label encoding function}, our method can effectively overcome the drawbacks of randomly initialized detection heads, leading to more stable optimization and better performance.
	\item Our method is demonstrated to be generally applicable in most of modern object detection systems. Compared with previous methods like \cite{mimickingdetectionkd,dcnv2,shen2017dsod}, though various auxiliary losses are also introduced, usually those methods rely on specified backbone architectures or detection frameworks. Furthermore, though the underlying formulations appear to be somewhat complex, the implementation of our approach is relatively simple.
\end{itemize}

\section{Related Work}
\label{sec:related_work}

\subsubsection{Auxiliary supervision.}

\emph{Auxiliary Supervision} is a common technique to improve the performance of the model in indirect ways, e.g., \emph{weight decay}, \emph{Center Loss} \cite{centerloss}, etc. Among various auxiliary supervision methods, \emph{Multi-task Learning} (MTL) \cite{MTL} methods are used commonly. MTL solves multiple tasks using a single model. By sharing parameters between the tasks, inductive bias is transferred and better generalization is gained. In object detection, \emph{Mask R-CNN} \cite{maskrcnn} combines object detection with instance segmentation by adding a simple mask branch to \emph{Faster R-CNN} model \cite{fasterrcnn}. The MTL strategy can improve the performance of the detection branch efficiently, but it requires additional mask annotation. \cite{recyclingannotations}, on the contrary, does not need additional annotations, but it requires carefully-designed auxiliary tasks.

\emph{Deeply Supervise} is another common method of auxiliary supervise. Instead of introducing additional tasks, Deeply Supervise introduces supervision on additional layers. \emph{DSN} \cite{deeplysupervise} first proposes the concept by adding additional supervision on the hidden layers. \emph{Inception} \cite{inception} also uses similar auxiliary classifiers on lower stages of the network. In semantic segmentation, \emph{PSPNet} \cite{pspnet} and \emph{ExFuse} \cite{exfuse} adopt Deeply Supervise in order to improve the low-level features. In object detection, \emph{DSOD} \cite{shen2017dsod} utilizes Deeply Supervise with dense connections to enable from-scratch training. In our method, we adopt the idea of Deeply Supervise by proposing a label encoding function, with which we can map the labels into latent embedding for auxiliary intermediate supervision.

\subsubsection{Knowledge distillation.}

our method shares some common inspiration with \emph{Knowledge Distillation} \cite{knowledgedistillation,fitnets,attentiondistillation,kdgan}. In Knowledge Distillation, the training is a two-step process. A large teacher model is trained first. Then its predictions are used to supervise a smaller student model. Knowledge Distillation has been used in several fields, e.g., face \cite{kdface}, speech \cite{kdspeech}, re-id \cite{darkrank}. There are several works focusing on object detection as well: \cite{chandrakerdetectionkd} uses balanced loss on classification, bounded loss on regression and $L2$ loss on feature; \cite{dcnv2} and \cite{mimickingdetectionkd} propose their distillation methods based on RoIs; \cite{finegraineddetectionkd} so that distillation focuses on object-local areas.

From the distillation perspective, the \emph{label encoding function} is the teacher model in our pipeline. It is trained in the first step and utilized for supervision in Step 2. But it is a relatively simple architecture and does not involve feature in real world. On the contrary, traditional distillation models rely heavily on a big teacher model. Usually, the stronger the teacher model is, the better distillation performance it can give. However, teacher models with high performance are not always available in practice. The state-of-the-art models are the best teachers we can find. This limits the performance of traditional distillation.

\subsubsection{Label encoding.}

There are several works that use label encoding to boost training \cite{bengio2010label,akata2013label,sun2017label,xie2016top}. However, few evaluate in supervised object detection task. Among them, our method is most similar to \cite{AEseg}. \cite{AEseg} uses an AutoEncoder to model the labels of semantic segmentation. The AutoEncoder is then used to perform auxiliary supervision. Compared with our method, there are two main differences: first, in object detection, label structures hardly exist. Segmentation has rich information in label structures thanks to the outline of regions in annotation, e.g. a cat has a long tail, a thick body and a small head. Whereas in object detection, such structures are very limited, since all objects are just boxes with different scales and aspect ratios. Second, we propose a joint optimization scheme that introduces auxiliary structures for training AutoEncoder, which we empirically find vital to the performance. Whereas in \cite{AEseg}, the AutoEncoder is trained independently.

\section{Method}
\label{sec:method}

\subsection{Intermediate Auxiliary Supervision}
\label{sec:supervision}

As mentioned in the introduction, the core of our method is to define the supervision term $\mathcal{R}(\cdot, \cdot)$ in Eq.~\ref{eq:ours}, which is expected to provide feasible supervision to the backbone training. Intuitively, the auxiliary loss should encourage the latent feature generated by the backbone network to be close to some ``ideal'' embedding $\mathcal{T}(I, y)$ for each training sample:
\begin{equation}
	\mathcal{R}(f(I; \theta_f), y) = \mathcal{L}_{dis} (f(I; \theta_f), \mathcal{T}(I, y)) ,
\end{equation}
where $\mathcal{L}_{dis}(\cdot, \cdot)$ represents the distance measurement. 
Therefore, a problem rises: how to define the so-called ``ideal'' feature $\mathcal{T}(I, y)$? Obviously, the calculation of $\mathcal{T}(I, y)$ \textbf{cannot} directly rely on the training of the detection head $d(\cdot; \theta_d)$, otherwise it may be unstable and redundant to the existing detection loss $\mathcal{L}_{det}$. 

Let us think for a further step. If we have finished the optimization in Eq.~\ref{eq:ours} via some way, i.e. the corresponding optimal weights $\theta_f^*$ and $\theta_d^*$ have been obtained, we can intuitively define the inverse of the detection head $d^{-1}(y; \theta_d^*)$ as the ``optimal'' intermediate embedding. So,
\begin{equation}
	\mathcal{R}(f(I; \theta_f), y) = \mathcal{L}_{dis}( f(I; \theta_f), d^{-1}(y; \theta_d^*) ).
\label{eq:inverse}
\end{equation}
We argue that the definition of $\mathcal{R}(\cdot,\cdot)$ is feasible because if the auxiliary loss tends to zero, it is easy to verify that the detector will predict the ground truth $y$ exactly. Unfortunately, Eq.~\ref{eq:inverse} cannot be directly used in the optimization. \textbf{First}, to substitute Eq.~\ref{eq:inverse} into Eq.~\ref{eq:ours}, we find $\theta_d^*$ exists in both side of the equation -- we cannot determine the value in advance. \textbf{Second}, even though $\theta_d^*$ is given, the inverse form $d^{-1}(\cdot; \theta_d^*)$ is still difficult to be calculated due to the high nonlinearity of neural networks (actually the inverse is generally not unique). 

We deal with the second problem firstly. Notice that for any $y$, we have $d^{-1} \circ d (y; \theta_d^*) \equiv y $. Motivated by this, to approximate $d^{-1}(\cdot; \theta_d^*)$ we introduce a new network $h(\cdot; \psi)$, whose parameters are learned by the optimization:
\begin{equation}
	\psi^* = \mathop{\arg\min}_\psi \mathbb{E}_{(I, y)\sim \mathcal{D}} \quad 
	\mathcal{L}_{det} ( d( h(y; \psi); \theta_d^*), y ) .
\label{eq:ae}
\end{equation}
Here, $\mathcal{L}_{det}(\cdot, \cdot)$ is the detection loss, following the definition in Eq.~\ref{eq:det}. Intuitively, $h(\cdot; \psi^*)$ maps the ground truth label $y$ into the latent feature space and $d(\cdot; \theta_d^*)$ recovers the label from the latent representation. So, we say that $h(\cdot; \psi^*)$ approximates the ``inverse'' of $d(\cdot; \theta_d^*)$. It is worth noting that the composite function $(h\circ d)(\cdot; \psi^*, \theta_d^*)$ actually represents an \textbf{AutoEncoder} defined in the \emph{label space}. Thus we name $h(\cdot)$ as \emph{label encoding function}. Thanks to the approximation, we rewrite Eq.~\ref{eq:inverse} as follows:
\begin{equation}
	\mathcal{R}(f(I; \theta_f), y) = \mathcal{L}_{dis} (f(I; \theta_f), h(y; \psi^*) ).
\label{eq:sub}
\end{equation}

Then we come back to the first problem. In Eq.~\ref{eq:sub}, note that the optimization of $\psi^*$ still implies $\theta_d^*$ (Eq.~\ref{eq:ae}). So, in our formulations (Eq.~\ref{eq:ours}, \ref{eq:sub} and \ref{eq:ae}) there still exists the recursive dependence on $\theta_d^*$. To get out of the dilemma, we use an \emph{unrolling} trick, i.e. recursively substituting Eq.~\ref{eq:sub} and Eq.~\ref{eq:ae} into Eq.~\ref{eq:ours}. Thus we obtain the final formulations (please refer to the appendix for the detailed derivation):
\begin{equation}
	\theta_f^*, \theta_d^* = \mathop{\arg\min}_{\theta_f, \theta_d} {}\mathbb{E}_{(I, y)\sim\mathcal{D}} \quad
	\mathcal{L}_{det}\left(d( f(I; \theta_f); \theta_d), y \right)
	+ \lambda  \mathcal{L}_{dis}( f(I; \theta_f), h(y; \psi^*) ), 
	\label{eq:final_sup}
\end{equation}
where
\begin{equation}
\begin{split}
	\psi^* = & \mathop{\arg\min}_\psi \mathbb{E}_{(I, y)\sim \mathcal{D}} \quad 
		\mathcal{L}_{det} ( d( h(y; \psi); \hat{\theta}_d), y ),   \\
		s.t. \qquad
		\hat{\theta}_d=&\mathop{\arg\min}_{\theta'_d} \left[ \min_{\theta'_f} 
		\mathbb{E}_{(I, y)\sim\mathcal{D}} 
		\mathcal{L}_{det}(\theta'_f, \theta'_d)
		+ \lambda  \mathcal{L}_{dis} (f(I; \theta'_f), h(y; \psi) )
		\right]. 
\label{eq:final_ae}
\end{split}
\end{equation}
Here $\mathcal{L}_{det}(\theta'_f, \theta'_d)$ is short for $\mathcal{L}_{det}(d( f(I; \theta'_f); \theta'_d), y )$.

Eq.~\ref{eq:final_sup} and Eq.~\ref{eq:final_ae} compose the core idea of our method. The formulations actually imply a \textbf{two-step} training pipeline. In the first step, by optimizing the auxiliary \emph{AutoEncoder} defined in Eq.~\ref{eq:final_ae}, we obtain an encoding function $h(\cdot; \psi^*)$ mapping the ground-truth label map $y$ into the latent space. Then in the second step, we train the detection framework with the intermediate supervision of $h(y; \psi^*)$, as described in Eq.~\ref{eq:final_sup}. In the next subsections, we will introduce the optimization details. 

\subsection{Step 1: AutoEncoder Training}
\label{sec:autoencoder}

In this subsection we aim to derive the \emph{label encoding function} $h(\cdot; \psi^*)$ via Eq.~\ref{eq:final_ae}. However, directly solving Eq.~\ref{eq:final_ae} is not easy -- since $\psi$ exists in both the target and the constraint, it is actually a \emph{bilevel optimization} problem, which seems nontrivial to be implemented with current deep learning tools. Therefore, we propose to relax the formulation into \emph{joint optimization} scheme, as follows:
\begin{equation}
\begin{split}
	\psi^*, \hat{\theta}_d &= \mathop{\arg\min}_{\psi, \theta'_d}  \min_{\theta'_f} 
	\mathbb{E}_{(I, y)\sim\mathcal{D}} \quad
			\mathcal{L}_{det} ( d(h (y; \psi); \theta'_d), y )  \\
			&+ \lambda_1 \mathcal{L}_{det}(d( f(I; \theta'_f); \theta'_d), y )
			+ \lambda_2  \mathcal{L}_{dis}( f(I; \theta'_f), h(y; \psi) ),
\label{eq:relax}
\end{split}
\end{equation}
where $\lambda_1$ and $\lambda_2$ are balanced coefficients, while in our experiment we just trivially set them to 1. It is clear that Eq.~\ref{eq:relax} simply corresponds to a multi-task training paradigm with three loss terms: the first one is \emph{reconstruction loss} (\textbf{L1}) for the label's AutoEncoder; the second term is the common detection loss (\textbf{L2}), which enforces $d(\cdot; \theta'_d)$ to be a \emph{valid} detection head; the third loss (\textbf{L3}) minimizes the gap between the two latent spaces (namely the outputs of the backbone $f(\cdot; \theta'_f)$ and label encoding function $h(\cdot; \psi)$ respectively). 

\begin{figure}
	\centering
	\includegraphics[width=0.9\textwidth]{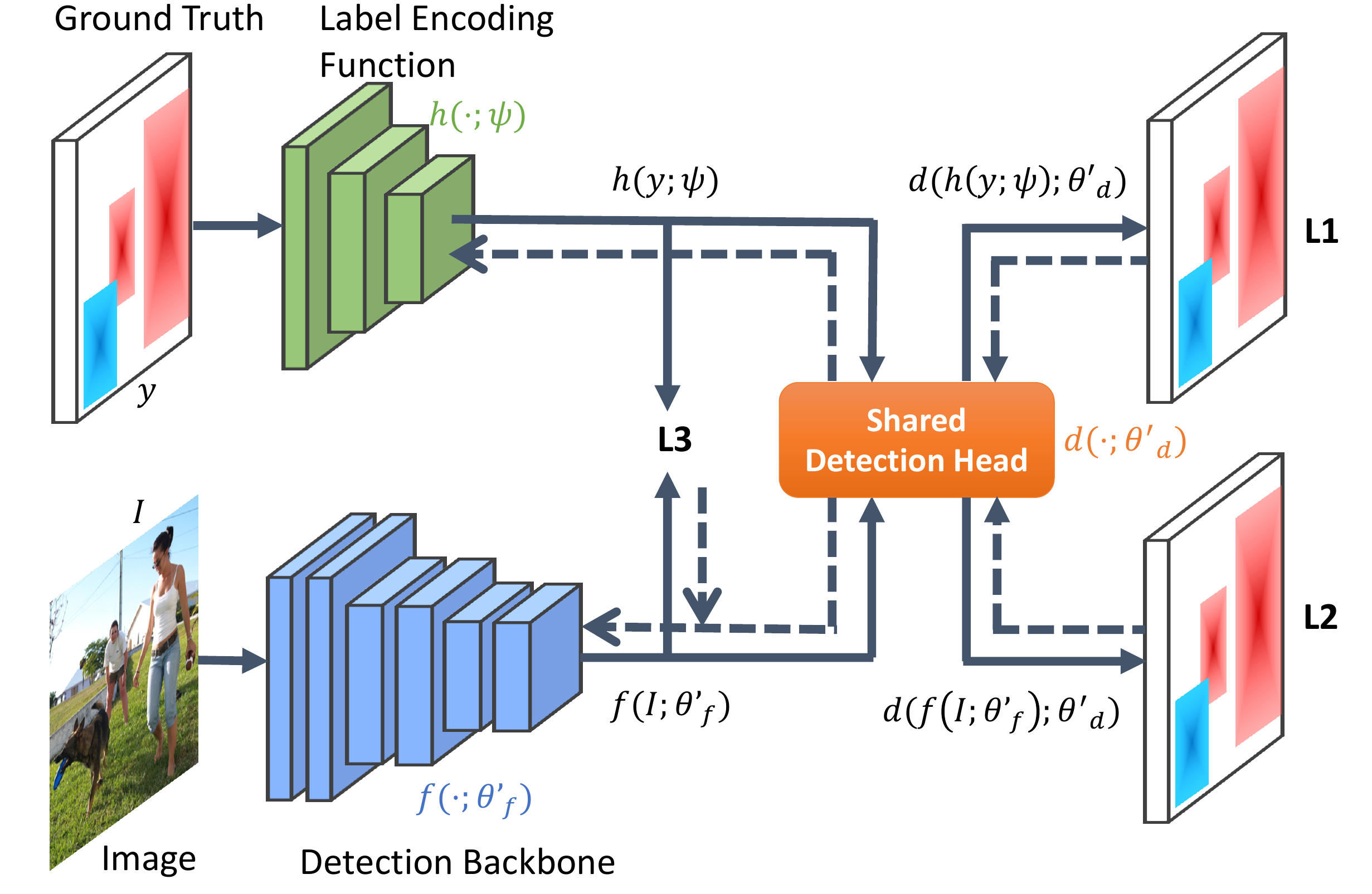}
	\caption{Step 1: AutoEncoder training. \textbf{L1} -- AutoEncoder reconstruction loss; \textbf{L2} -- detection loss; \textbf{L3} -- distance minimization loss; please refer to Eq.~\ref{eq:relax} for details. The solid and dashed lines indicate the forward and backward flows respectively }
	\label{fig:autoencoder}
\end{figure}

Fig.~\ref{fig:autoencoder} illustrates the implementation and optimization of Eq.~\ref{eq:relax}. According to Eq.~\ref{eq:relax}, the same detection head $d(\cdot; \theta'_d)$ is applied in both \textbf{L1} and \textbf{L2} terms -- which is why we mark ``shared detection head'' in Fig.~\ref{fig:autoencoder}. It is also worth noting that we forbid the gradient flow from \textbf{L3} to the label encoding function $h(\cdot; \psi)$. The motivation is, in Eq.~\ref{eq:final_ae} (which is the original form of Eq.~\ref{eq:relax}), the optimization of $\theta'_d$ does not directly affect $\psi$, thus we follow the property in the implementation. We empirically find the above details are critical to improve the final performance. 

\paragraph{Initialization.} Before optimization, we follows the common practice of initialization method, i.e. using pretrained weights (e.g. pretrained on \emph{ImageNet} \cite{imagenet}) for backbone parameters $\theta'_f$ and Gaussian random weights for $\psi$ and $\theta'_d$. One may argue that according to the introduction, randomly initialized detection head $d(\cdot; \theta'_d)$ may cause unstable training. But actually, since this training step mainly aims to learn the label encoding function $h(\cdot; \psi)$, the detection backbone $f(\cdot; \theta'_f)$ and the detection head $d(\cdot; \theta'_d)$ are thus ``auxiliary structures'' in this step, whose performances are not that important. Furthermore, as we will introduce, the architecture of $h(\cdot; \psi)$ is relatively simple, so the optimization seems not difficult.

\subsection{Step 2: Detector Training with Intermediate Supervision}
\label{sec:detector}

\begin{figure}
	\centering
	\includegraphics[width=0.9\textwidth]{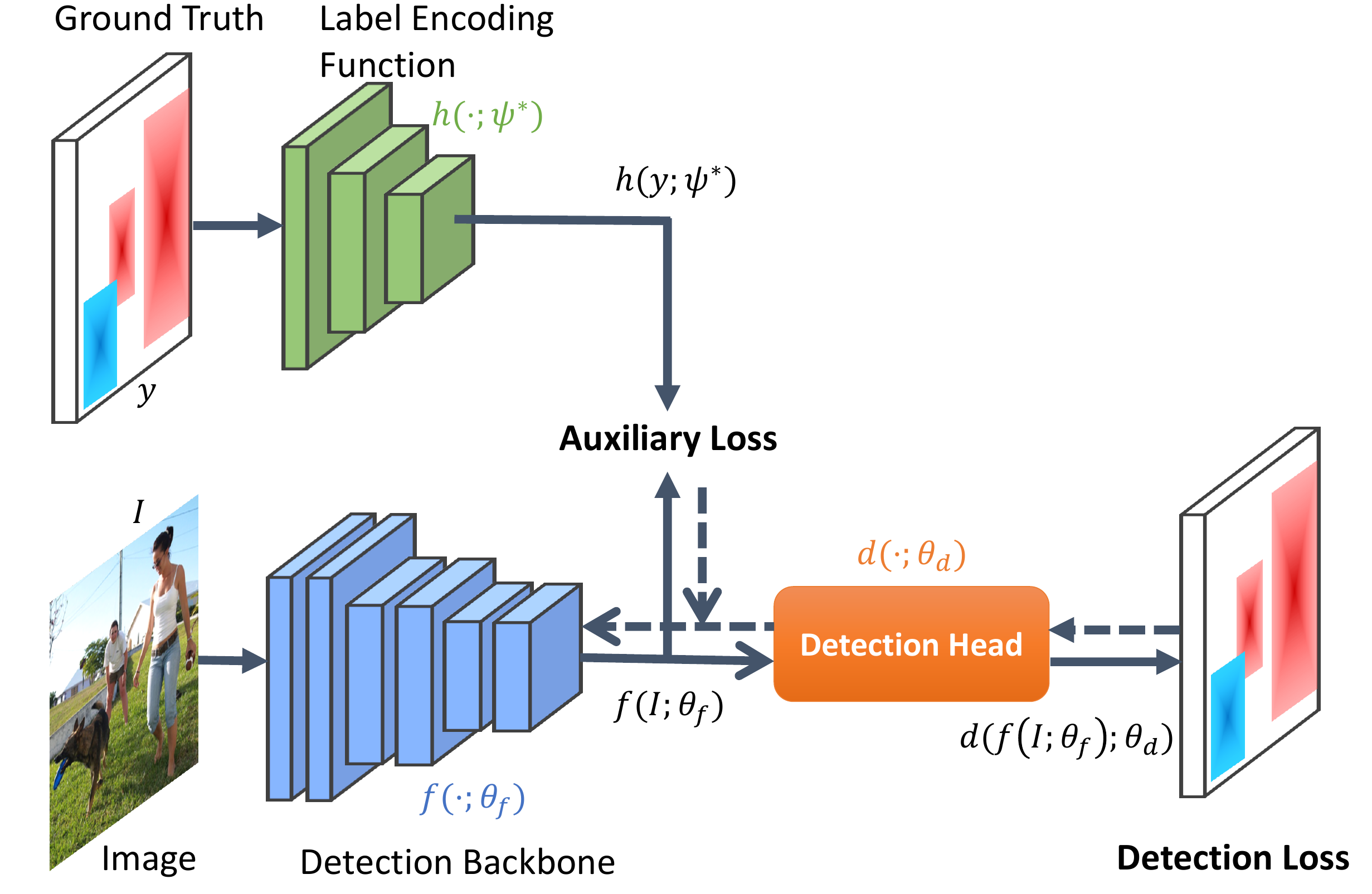}
	\caption{Step 2: Detector training with intermediate supervision. Please refer to Eq.~\ref{eq:final_sup} for the detailed definitions. The solid and dashed lines indicate the forward and backward flows respectively }
	\label{fig:supervise_decoder}
\end{figure}

After the \emph{label encoding function} $h(\cdot; \psi^*)$ has been learned, we then use it as the intermediate supervision to improve object detector training, according to Eq.~\ref{eq:final_sup}. Fig.~\ref{fig:supervise_decoder} illustrates the implementation. In addition to the common detection loss, we introduce an auxiliary loss $ \lambda \mathcal{L}_{dis}( f(I; \theta_f), h(y; \psi^*) )$ to \emph{directly} supervise the detection backbone. The coefficient $\lambda$ is also trivially set to 1. Besides, Eq.~\ref{eq:final_sup} also suggests that $\psi^*$ is fixed rather than optimization variable. So, we block the gradient flow from the auxiliary loss to $h(\cdot; \psi^*)$, as shown in the figure. After training, the auxiliary structure -- $h(\cdot; \psi^*)$ -- is then removed. The resulted $(f \circ d) (\cdot; \theta_f^*, \theta_d^*)$ is the learned object detector we expected. 

Another important detail on the implementation is \emph{initialization}. From Eq.~\ref{eq:final_sup} and Eq.~\ref{eq:final_ae} we know that in the two training steps, the detection backbones $f(\cdot)$ and the detection heads $d(\cdot)$ shares the same network architecture respectively, however, whose parameters are not necessarily the same. So, in Step 2, we \emph{reinitialize} the the backbone parameters $\theta_f$ (using \emph{ImageNet} pretrained weights, for instance) before training. As for the detection head parameters $\theta_d$, empirically we find that initializing them with the corresponding parameters $\hat{\theta}_d$ learned in Step 1 (see Eq.~\ref{eq:relax}) results in better performance and stable convergence. It may be because the pretrained detection head can provide less gradient noise to the backbone, compared with the randomly initialized heads.

\subsection{Implementation Details and Remarks }
\label{sec:details}

\subsubsection{Ground-truth label representation. } 

As mentioned above, in both two training steps the \emph{label encoding function} $h(\cdot)$ needs to take ground-truth labels $y$ as the network inputs. It is nontrivial because in detection task, each image contains different numbers of instances, each of which may have various class labels and bounding boxes. We have to produce a fixed-length label map that contains all the ground-truth information for each image. 

We propose to use a $C\times H \times W$ tensor to represent the ground-truth objects in one image, where $H\times  W$ equals to the image size and $C$ is the number of classes in the dataset (e.g. 80 for \emph{COCO} \cite{coco} dataset). For an object of the $c$-th class, we fill the corresponding region (according to the bounding box) in the $c$-th channel with positive values: the value ranges from 1 at the object center to 0.5 in the box boundary, which decays linearly. Fig.~\ref{fig:autoencoder} and Fig.~\ref{fig:supervise_decoder} visualize the encoding. Specially, if two bounding boxes of the same class overlap with each other, the joint region is filled with larger values of those calculated separately. Additionally, in training, the boxes are augmented by multiplying a random number between 0 and 1 with a probability of 0.5. Other values in the tensor remain to be zeros.

\subsubsection{Architecture of label encoding function.} 

For ease of optimization, we use relatively simple architecture to implement $h(\cdot)$. The design of the structure is inspired by \emph{ResNet} \cite{resnet}, while the number of residual blocks in each stage reduces to $\{1, 2, 2, 1\}$ respectively. In addition, the \emph{Max Pooling} layer is replaced by stride convolution. The input channels is set to 80 to satisfy the number of classes in \emph{COCO} \cite{coco} dataset. \emph{Batch Normalization} \cite{batchnorm} is \textbf{not} used here. We use the same architecture for all experiments in the paper. Please refer to the appendix for details. 

\subsubsection{Multi-scale intermediate supervision. } 

Recently, state-of-the-art detection frameworks like \cite{fpn,retinanet,fcos}  usually introduce \emph{Feature Pyramid Networks} (FPNs) to generate multi-scale feature maps, which greatly improves the capacity to detect objects of various sizes. Our approach can be easily generalized to multi-scale cases. First, we attach one \emph{FPN} structure to the label encoding function $h(\cdot)$ so that it can produce multi-resolution representations. Then in both Step 1 and Step 2,  we make the intermediate supervision terms $\mathcal{L}_{dis}(f(I), h(y))$ (see Eq.~\ref{eq:relax} and Eq.~\ref{eq:final_sup}) applied on all the scale levels. As shown in the following experiments, our method can effectively boost the detection frameworks with FPNs. 

\subsubsection{Distance measurement. }

In Eq.~\ref{eq:final_sup} and Eq.~\ref{eq:relax}, the \emph{distance measurement} term $\mathcal{L}_{dis}(\cdot, \cdot)$ is used to minimize the difference between two feature maps. One simple alternative is to use $L2$-distance directly. However, there are several issues as follows: 1) the sizes of the two feature maps may be different; 2) since the feature maps are generated from different domains respectively, directly minimizing their difference may suffer from very large gradient. So, we propose to introduce a \emph{feature adaption} block into the distance measurement, which is defined as follows:
\begin{equation}
    \mathcal{L}_{dis}(\mathbf{x}_f, \mathbf{x}_h)
    \triangleq
    \min_\phi \left\|\text{LN}(\mathcal{A}(\mathbf{x}_f; \phi)) - \text{LN}(\mathbf{x}_h)\right\|^2,
\end{equation}
where $\text{LN}(\cdot)$ means \emph{Layer Normalization} \cite{layernorm}; $\left\| \cdot \right\|$ is $L2$-distance; $\mathbf{x}_f$ and $\mathbf{x}_h$ are feature maps derived from the backbone and the \emph{label encoding function} respectively. $\mathcal{A}(\cdot)$ represents \emph{feature adaption network}, which acts as the transformer between the two domains. We implement $\mathcal{A}(\cdot)$ with three convolution layers, whose kernel size is $3\times 3$ and number of channels is 256. The parameters $\phi$ are learned jointly with the outer optimization. Similar to $h(\cdot)$, $\mathcal{A}(\cdot)$ is also an auxiliary structure thus will be discarded after training.  

\section{Experiment}
\label{sec:exp}

\subsection{Setup}

All our experiments are done with \emph{PyTorch} \cite{paszke2019pytorch}. We use \emph{COCO} \cite{coco} dataset to evaluate our method. Following the common practice \cite{fpn,retinanet}, we train our models with the union of 80k train images and a subset of 35k validation images (\emph{trainval35k}). We test our models in the rest 5k of validation images (\emph{minival}). All results are evaluated with mmAP, i.e. mAP@[0.5,0.95], using common single-scale test protocol. For both training and inference, we resize each images to 800 pixels on the shorter edge. The training batch size is a total of 16 in 8 GPUs. We mainly use so-called 1$\times$ schedule for training, which refers to 90k iterations with two learning rate decays at 60k and 80k iteration. 
We use almost the same training protocol for our Step 1 and Step 2 training, as well as all the counterpart baseline models respectively, with two exceptions: for Step 1, we find that adding L3 from the beginning cause L3 to be nearly zero. The network somehow finds a way to cheat, causing terrible results. So we add an additional 30k warmup iterations without L3, which we find sufficient to solve the problem; for Step 2, we remove the auxiliary loss in the last 10k iterations, which results in minor improvements. 
Since our training pipeline involves two steps, the total number of the iterations thus doubles. For fair comparison, we provide 2$\times$ schedule for baseline models as well, which refers to 180k iterations with two learning rate decays at 120k and 160k iteration.

\subsection{Main Results}

\setlength{\tabcolsep}{3pt}

\begin{table}[t]
\begin{center}
\caption{Experiments on various baselines (mmAP/\%)}
\label{tab:results}
\begin{tabular}{llccc}
\toprule
\textbf{Model} & \textbf{Backbone} & \textbf{Baseline(1x)} & \textbf{Baseline(2x)} & \textbf{Ours} \\ \midrule
\multirow{3}{*}{
\begin{tabular}{@{}l@{}}RetinaNet \cite{retinanet} \\ (\emph{our impl.})\end{tabular}
} & ResNet50 & 36.4 & 36.4 & \textbf{38.6} \\
 & ResNet101 & 38.3 & 38.5 & \textbf{40.5} \\
 & Res101-DCN & 41.1 & 41.5 & \textbf{42.9} \\
 \midrule
\multirow{3}{*}{
\begin{tabular}{@{}l@{}}FCOS \cite{fcos} \\ (\emph{our impl.})\end{tabular}
} & ResNet50 & 38.7 & 38.4 & \textbf{41.1} \\
 & ResNet101 & 41.0 & 40.5 & \textbf{43.1} \\
 & Res101-DCN & 43.7 & 43.5 & \textbf{45.8} \\
 \midrule
\multirow{3}{*}{
\begin{tabular}{@{}l@{}}FPN \cite{fpn} \\ (\emph{our impl.})\end{tabular}
} & ResNet50 & 36.8 & 37.6 & \textbf{38.6} \\
 & ResNet101 & 39.1 & 39.6 & \textbf{41.0} \\
 & Res101-DCN & 42.1 & 42.8 & \textbf{43.3} \\
\bottomrule
\end{tabular}
\end{center}
\vspace{-25pt}
\end{table}

In order to show the effectiveness of our model on different detection frameworks, we evaluate our method on \emph{RetinaNet} \cite{retinanet}, \emph{FCOS} \cite{fcos} and \emph{FPN} \cite{fpn}, which are representative baselines of one-stage detectors, anchor-free methods and two-stage frameworks respectively. For FCOS, we use ``improvements" in \cite{fcos}. We evaluate our method on various commonly-used backbones, including \emph{ResNet}-50, \emph{ResNet}-101 \cite{resnet} and \emph{Deformable Convolutional Networks} (DCNs) \cite{dai2017deformable}. 

Results are presented in Table \ref{tab:results}. Compared with the counterparts with 1$\times$ schedule, our method achieves performance gains of over 2\% on both ResNet-50 and ResNet-101 backbones. On ResNet-101-DCN, there are still relative improvements of $\sim$1.7\% in average. Compared with the baselines of 2$\times$ schedule, the gap remains considerable, which suggests that our improvements are not mainly brought by more training iterations. 
It is worth noting that although our training pipeline doubles the total number of iterations, we argue that our \emph{label encoding function} can usually be reused among different backbones (see the next subsection). Therefore in practice, we usually only need to run Step 1 only once for different models. 

\subsection{Ablation Study}

\subsubsection{Step 1: is joint optimization required?}

In Sec. \ref{sec:autoencoder}, to optimize Eq.~\ref{eq:relax} we propose a \emph{joint optimization} scheme to take all the three loss terms into account. Recall that in Step 1, only the learned \emph{label encoding function} will be reserved into the next stage. As a result, one may argue that whether the auxiliary structure, i.e. the detection backbone, is really necessary in training. In other words, the question is, can we only use \textbf{L1} (\emph{AutoEncoder reconstruction loss}) in Eq.~\ref{eq:relax} for this step? If it is true, the training step can be further simplified. Unfortunately, we find it not the case. 

To validate the argument, we conduct a comparison by removing \textbf{L2} and \textbf{L3} in Eq.~\ref{eq:relax} to derive the \emph{label encoding function}. Other settings such as Step 2 keep unchanged. The results are listed in Table~\ref{tab:pipelines}, while the modified counterparts are marked with ``reconstruction loss only''. We compare them on RetinaNet with ResNet-50 and ResNet-101 backbones. It is clear that, without the auxiliary backbone, our method (although still outperforms baseline models) shows significant degradation in precision.

\setlength{\tabcolsep}{6pt}
\begin{table}[t]
    \begin{center}
    \caption{Ablation study of removing the auxiliary structures in Step 1}
    \label{tab:pipelines}
    \begin{tabular}{llc}
    \toprule
    \textbf{Backbone}       & \textbf{Methods}      & \textbf{mmAP (\%)}   \\ \midrule
    \multirow{4}{*}{ResNet50}  & Baseline (1$\times$) & 36.4 \\
                            & Baseline (2$\times$) & 36.4 \\
                            & Ours (reconstruction loss only) & 37.6 \\
                            & Ours & \textbf{38.6} \\
    \midrule
    \multirow{4}{*}{ResNet101} & Baseline (1$\times$) & 38.3 \\
                            & Baseline (2$\times$) & 38.5 \\
                            & Ours (reconstruction loss only) & 39.7 \\
                            & Ours & \textbf{40.5} \\
    \bottomrule
    \end{tabular}
    \end{center}
\end{table}
    
\paragraph{Discussion and remarks. } In Step 1, although the existence of auxiliary structures is vital, we find the exact weights in the backbone are actually less important. From Eq.~\ref{eq:final_sup} and Eq.~\ref{eq:final_ae}, we know that $\theta'_f$ does not affect the optimization of $\theta_f$ directly. It only contributes to the optimization of $\psi^*$. Also, unlike $\theta_d$ which is inherited from $\hat{\theta}_d$ for initialization, $\theta_f$ is reinitialized exactly in Step 2. Therefore, the trained auxiliary detection backbone $f(\cdot; \theta'_f)$ in Step 1 is completely discarded. 

The observation inspires an interesting assumption: is the final performance actually insensitive to the detailed backbone architecture in Step 1? We try to verify the guess by using different backbones in Step 1 and Step 2. As reported in Table~\ref{tab:auxiliary_detector}, we use ResNet-50 as the auxiliary backbone in the first step. Whereas in Step 2, the final detection backbone is ResNet-101. Compared with the model whose backbones in both stages are ResNet-101, the performances almost keep the same. 
The new finding thus suggests another advantage of our method in practice. The \emph{label encoding function} $h(\cdot; \psi^*)$ can be \textbf{pretrained once but re-used for multiple detectors with different backbones}, as long as they have the same detection head. 
This property of our method greatly reduces the cost of the practical applications.

\begin{table}[t]
    \begin{center}
    \caption{Comparisons of different detection backbones in Step 1}
    \label{tab:auxiliary_detector}
    \begin{tabular}{ccc}
    \toprule
    \textbf{Step 1 Backbone} & \textbf{Step 2 Backbone} & \textbf{mmAP (\%)} \\
    \midrule
    ResNet50 & \multirow{2}{*}{ResNet101} & 40.5 \\
    ResNet101 & & \textbf{40.6} \\
    \bottomrule
    \end{tabular}
    \end{center}
    \vspace{-25pt}
    \end{table}
    
\subsubsection{Is Step 1 alone sufficient?}

In Step 1, we only aim to solve the \emph{label encoding function} for later intermediate supervision. However, the training framework in Step 1 is quite similar to that in Step 2, and there is a detection model (the auxiliary structure) that can be proceeded for testing. Intuitively, the detection model in Step 1 should improve as well. One may even guess that Step 1 alone is sufficient. We show the ablation in Table~\ref{tab:step1only}. We only use Step 1 and test the performance of the detection model (the auxiliary structure). We compare them on multiple models with ResNet50 backbone. Step1-only can indeed improve the detection model over baseline, but clearly it alone is not sufficient.

\begin{table}[t]
    \begin{center}
    \caption{Results of only using Step 1}
    \label{tab:step1only}
    \begin{tabular}{lllc}
    \toprule
    \textbf{Model} & \textbf{Backbone} & \textbf{Method} & \textbf{mmAP (\%)} \\
    \midrule
    \multirow{2}{*}{RetinaNet} & \multirow{2}{*}{ResNet50} & Step1-only & 37.5 \\
     & & Ours & \textbf{38.6} \\
    \midrule
    \multirow{2}{*}{FCOS} & \multirow{2}{*}{ResNet50} & Step1-only & 40.3 \\
     & & Ours & \textbf{41.1} \\
    \midrule
    \multirow{2}{*}{FPN} & \multirow{2}{*}{ResNet50} & Step1-only & 38.1 \\
     & & Ours & \textbf{38.6} \\
    \bottomrule
    \end{tabular}
    \end{center}
    \vspace{-25pt}
    \end{table}
    
\subsubsection{Step 2: do intermediate supervision and initialization matter?}

In Step 2, we use two methods to facilitate the optimization, i.e. \emph{intermediate supervision} on the backbone as well as the \emph{initialization} of the detection head. In Table~\ref{tab:distillation_and_finetune}, we show the ablation studies on them. The baseline framework is \emph{RetinaNet} \cite{retinanet} with ResNet-50 backbone. We also make the combinational studies of the case that using reconstruction loss only in Step 1 (please refer to Table~\ref{tab:pipelines}). The results suggest that both methods contribute to the final performance. 

\begin{table}[h]
    \begin{center}
    \caption{Intermediate supervision and initialization in Step 2}
    \label{tab:distillation_and_finetune}
    \begin{tabular}{cccc}
    \toprule
    \textbf{Step 1} & \textbf{Supervision} & \textbf{Initialization} & \textbf{mmAP (\%)} \\
    \midrule
    ResNet50 Baseline & & & 36.4 \\
    \midrule
    \multirow{3}{*}{
    \begin{tabular}{c}Ours \\ (reconstruction loss only)\end{tabular}
    }
    & \checkmark & & 37.0 \\
     & & \checkmark & 36.8 \\
     & \checkmark & \checkmark & \textbf{37.6} \\
    \midrule
    \multirow{3}{*}{Ours} & \checkmark & & 37.2 \\
     & & \checkmark & 37.7 \\
     & \checkmark & \checkmark & \textbf{38.6} \\
    \bottomrule
    \end{tabular}
    \end{center}
\vspace{-25pt}
\end{table}

\subsection{Comparison with Knowledge Distillation}

Our two-step pipeline resembles \emph{Knowledge Distillation} (KD). Actually, if we train an object detector alone in Step 1 instead of our \emph{label encoding function} with a joint framework, and use it in Step~2 for supervision, the method becomes KD. In Table~\ref{tab:compare_kd} we show comparison between our method and the alternative mentioned above, denoted as ``Vanilla KD". On a lightweight backbone, i.e. MobileNet, our method can reach similar performance to Knowledge Distillation, although we only use a label encoding function instead of a heavy ResNet-50 that extracts ``real" features. On a heavier backbone, i.e. ResNet-50, our method outperforms KD with ResNet-50 and ResNet-101 as teachers, whose improvements are limited due to the small performance gap between teacher and student. Knowledge distillation requires a teacher network that is strong enough, which is usually not easy to find when the student network is already strong. Our method, on the other hand, is not limited by it.

\begin{table}[t]
    \begin{center}
    \caption{Comparison with Knowledge Distillation (\%) }
    \label{tab:compare_kd}
    \begin{tabular}{cccc}
    \toprule
    \textbf{Backbone} & \textbf{Method} & \textbf{Teacher Network} & \textbf{mmAP} \\
    \midrule
    \multirow{3}{*}{MobileNet} & Baseline & - & 27.7 \\
     & Vanilla KD & ResNet50 & 29.7 \\
     & Ours & Label Encoding Function & \textbf{29.8} \\
    \midrule
    \multirow{4}{*}{ResNet50} & Baseline & - & 36.4 \\
     & Vanilla KD & ResNet50 & 37.3 \\
     & Vanilla KD & ResNet101 & 37.8 \\
     & Ours & Label Encoding Function & \textbf{38.6} \\
    \bottomrule
    \end{tabular}
    \end{center}
    \vspace{-5pt}
    \end{table}

\setlength{\tabcolsep}{10pt}
\subsection{Performance on Mask Prediction}

Above we mainly focus on object detection. However, our previous discussion when proposing the method (Sec.~\ref{sec:intro} and Sec.~\ref{sec:method}) is based on the structure and optimization of detection networks, not object detection task itself. Thus it is likely that our method can be extended to other tasks with similar framework. We tested our method on Mask R-CNN \cite{maskrcnn}, which produces mask prediction in instance segmentation, but has a similar framework to FPN. It is worth noting that for Mask R-CNN, we use masks instead of boxes as the input for label encoding function. Results are presented in Table~\ref{tab:maskrcnn}. It indicates our method improves mask prediction as well.

\begin{table}[t]
\begin{center}
\caption{Experiments on MaskRCNN (mmAP/\%)}
\label{tab:maskrcnn}
\begin{tabular}{llcc}
\toprule
\textbf{Backbone} & \textbf{Method} & \textbf{box} & \textbf{mask} \\ \midrule
\multirow{3}{*}{ResNet50} & Baseline (1$\times$) & 37.4 & 34.2 \\
 & Baseline (2$\times$) & 38.2 & 34.6 \\
 & Ours & \textbf{39.1} & \textbf{35.6} \\
 \midrule
\multirow{3}{*}{ResNet101} & Baseline (1$\times$) & 40.0 & 36.0 \\
 & Baseline (2$\times$) & 40.6 & 36.4 \\
 & Ours & \textbf{41.7} & \textbf{37.6} \\
\bottomrule
\end{tabular}
\end{center}
\vspace{-25pt}
\end{table}




\section{Conclusions}

In this paper, we propose a new training pipeline for object detection systems. We design a feature encoding function and utilize it to introduce intermediate supervision on the detection backbone. Our method is generally applicable and efficient, adding no extra cost in inference time. To show its ability, we evaluate it on a variety of detection models and gain consistent improvement.

%
%
\bibliographystyle{splncs04}
\bibliography{egbib}

\appendix
\clearpage\section*{Appendix}

\section{Architecture of Label Encoding Function}

\setlength{\tabcolsep}{6pt}
\begin{table}[h]
    \centering
    \caption{Architecture of our label encoding function. It has 19 layers. Most stages have the same output channels and stride as in \emph{ResNet}-50 and \emph{ResNet}-101, which is convenient for later supervision. Except that the first convolution has 80 and 128 for input and output channels respectively, instead of 3 and 64, in order to satisfy \emph{COCO} dataset. We also remove the \emph{max pooling} and we do not use \emph{batch normalization} }
    \begin{tabular}{c|c|c|c|c}
        \textbf{Stage} & \textbf{Block} & \textbf{Kernel Size} & \textbf{Stride} & \textbf{Output Channels} \\
        \hline
        \hline
        Stage1 & Conv & 7 $\times$ 7 & 2 & 128 \\
        \hline
        
        \multirow{3}{*}{Stage2} & \multirow{3}{*}{ResBlock} & 1 $\times$ 1 & 1 & 64 \\
         & & 3 $\times$ 3 & 2 & 64 \\
         & & 1 $\times$ 1 & 1 & 256 \\
        \hline
        
        \multirow{6}{*}{Stage3} & \multirow{3}{*}{ResBlock} & 1 $\times$ 1 & 1 & 128 \\
         & & 3 $\times$ 3 & 2 & 128 \\
         & & 1 $\times$ 1 & 1 & 512 \\
        \cline{2-5}
         & \multirow{3}{*}{ResBlock} & 1 $\times$ 1 & 1 & 128 \\
         & & 3 $\times$ 3 & 1 & 128 \\
         & & 1 $\times$ 1 & 1 & 512 \\
        \hline
        
        \multirow{6}{*}{Stage4} & \multirow{3}{*}{ResBlock} & 1 $\times$ 1 & 1 & 256 \\
         & & 3 $\times$ 3 & 2 & 256 \\
         & & 1 $\times$ 1 & 1 & 1024 \\
        \cline{2-5}
         & \multirow{3}{*}{ResBlock} & 1 $\times$ 1 & 1 & 256 \\
         & & 3 $\times$ 3 & 1 & 256 \\
         & & 1 $\times$ 1 & 1 & 1024 \\
        \hline
        
        \multirow{3}{*}{Stage5} & \multirow{3}{*}{ResBlock} & 1 $\times$ 1 & 1 & 512 \\
         & & 3 $\times$ 3 & 2 & 512 \\
         & & 1 $\times$ 1 & 1 & 2048 \\
        \hline
    
    \end{tabular}
    \label{tab:my_label}
\vspace{-25pt}
\end{table}

\section{Derivation of Eq. (7, 8)}

In Sec. 1 and Sec. 3 we introduce our model as follows:
\begin{equation}
\theta_f^*, \theta_d^* = \mathop{\arg\min}_{\theta_f, \theta_d}
\mathbb{E}_{(I, y)\sim \mathcal{D}}
\mathcal{L}_{det}(d(f(I; \theta_f); \theta_d), y)
+ \lambda \mathcal{L}_{dis}(f(I; \theta_f), h(y; \psi^*)),
\label{apeq:det}
\end{equation}
where the optimal weights $\psi^*$ of the \emph{label encoding function} ($h(\cdot)$) is derived from:
\begin{equation}
\psi^* = \mathop{\arg\min}_{\psi} \mathbb{E}_{(I, y)\sim \mathcal{D}}
\mathcal{L}_{det}(d(h(y;\psi); \theta_d^*), y).
\label{apeq:ae}
\end{equation}
Clearly, there exist nested dependencies on the two variables $\theta_d^*$ and $\psi^*$. Thus the above equations are infeasible to compute directly. 

Notice that in Eq.~\ref{apeq:det}, $\psi^*$ actually acts as a \emph{constant} in the optimization. We define a function $\hat{\theta}_d(\cdot)$ as follows:
\begin{equation}
\hat{\theta}_d(\psi) \triangleq \mathop{\arg\min}_{\theta'_d}
\left[
\min_{\theta'_f} \mathbb{E}_{(I, y)\sim\mathcal{D}}
\mathcal{L}_{det}(d(f(I; \theta'_f)); \theta'_d), y)
+ \lambda \mathcal{L}_{dis}(f(I; \theta'_f), h(y; \psi))
\right].
\label{apeq:thetafun}
\end{equation}
Compared with Eq.~\ref{apeq:det}, we use $\theta'_d$, $\theta'_f$ instead of $\theta_d$ and $\theta_f$ respectively for distinguishing. Easy to find that $\hat{\theta}_d(\psi^*)=\theta_d^*$. Then, we can rewrite Eq.~\ref{apeq:ae} as follows:
\begin{equation}
\psi^*=\mathop{\arg\min}_{\psi} \mathcal{F(\psi, \psi^*)},
\label{apeq:recursive}
\end{equation}
where
\begin{equation}
\mathcal{F}(\psi, \psi^*) = \mathbb{E}_{(I, y)\sim\mathcal{D}} 
\mathcal{L}_{det}(d(h(y;\psi);\hat{\theta}_d(\psi^*)), y). 
\end{equation}
Eq.~\ref{apeq:recursive} suggests that we need to find a certain $\psi^*$ satisfying that the optimal point of the partial function $\mathcal{F}(\cdot, \psi^*)$ is also $\psi^*$, i.e. $\min_\psi \mathcal{F}(\psi, \psi^*)=\mathcal{F}(\psi^*, \psi^*)$. It motivates us to approximate $\psi^*$ with the following optimization, since Eq.~\ref{apeq:recursive} is nontrivial to compute directly:
\begin{equation}
\begin{split}
\psi^* &\simeq \mathop{\arg\min}_\psi \mathcal{F}(\psi, \psi) \\
&= \mathop{\arg\min}_\psi
\mathbb{E}_{(I, y)\sim\mathcal{D}} 
\mathcal{L}_{det}(d(h(y;\psi);\hat{\theta}_d(\psi)), y), \\
\end{split}
\end{equation}
which derives our formulations in the text.

\section{Feature Visualization}

\begin{figure}[h]
\centering
    \includegraphics[width=0.8\textwidth]{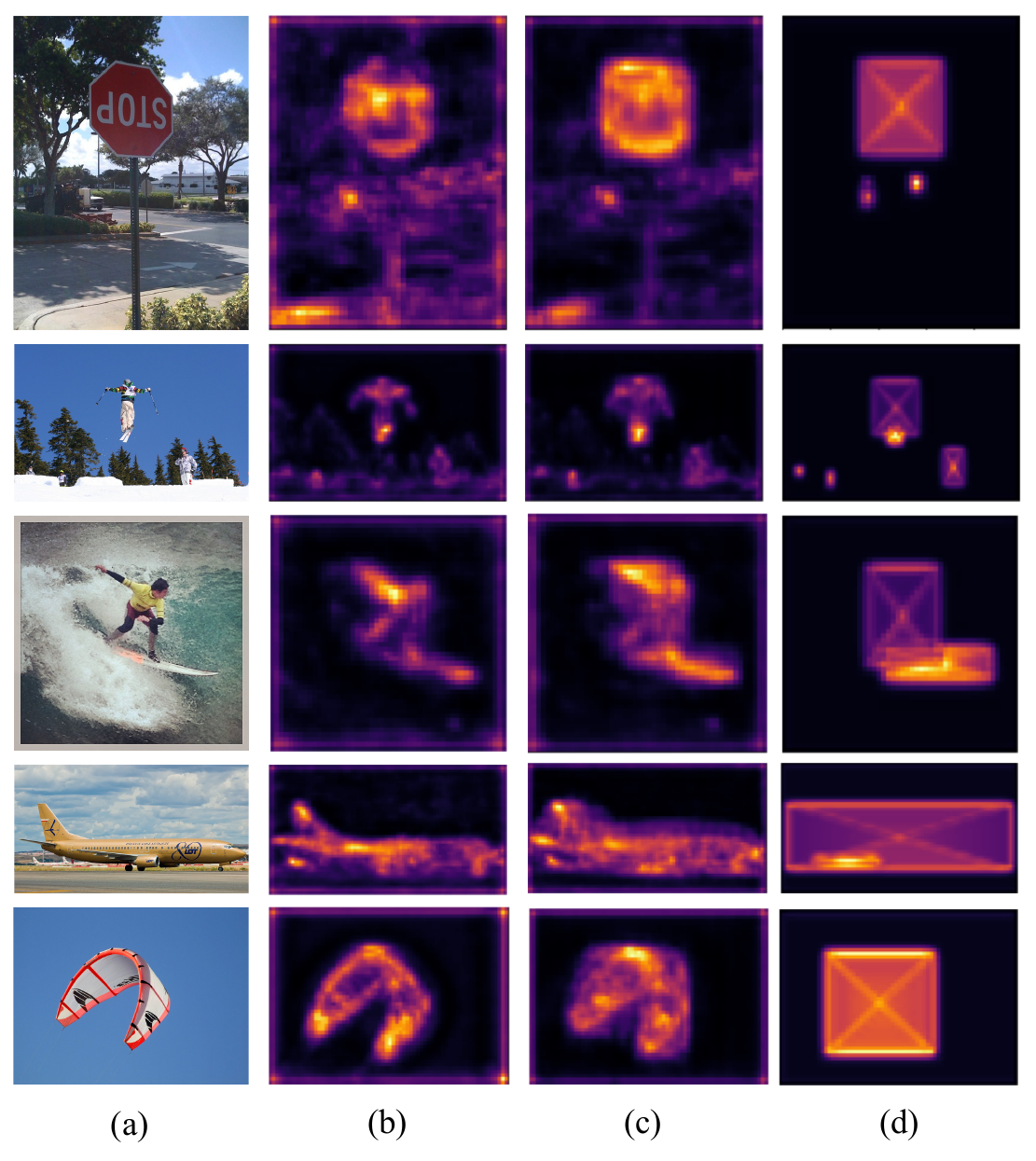}
    \caption{Visualization of feature in baseline and ours. (a) The original images. (b) Feature from baseline. (c) Feature from ours. (d) Feature from our encoding function }
    \label{fig:adaptation}
\end{figure}

In this section we analyze our method with visualization on feature maps. We pick the second layer of the multi-scale feature maps from RetinaNet. We use images in validation set. We visualize each feature map with its intensity. Specifically, we use the $L2$ norm of each pixel. The larger the $L2$ norm is (i.e., the stronger the intensity), the brighter it is in the figure. Visualization results are shown in Fig.~\ref{fig:adaptation}. Four columns are: (a) The original images. (b) Feature from baseline models. (c) Feature from our model. (d) Feature from our encoding function, which is the optimization target for (c). Compared with feature from baseline, which has clear boundary at the outline of each object, feature from ours is closer to boxes. Under the supervision from (d), the feature extends outside the object outline and ``tries to reach the box edges''. We believe this is beneficial for later instance extraction by detection head. Note that Fig.~\ref{fig:adaptation} is just a spatial projection of features. Information across channels is not visible here.

\end{document}